\documentclass[conference]{IEEEtran}
\usepackage{cite}
\usepackage{amsmath,amssymb,amsfonts}
\usepackage{algorithmic}
\usepackage{graphicx}
\usepackage{textcomp}
\usepackage{xcolor}
\def\BibTeX{{\rm B\kern-.05em{\sc i\kern-.025em b}\kern-.08em
    T\kern-.1667em\lower.7ex\hbox{E}\kern-.125emX}}

\usepackage{physics} 

\usepackage{glossaries} 

\usepackage[hyphens]{url}
\usepackage{hyperref}
\usepackage[hyphenbreaks]{breakurl}

\newacronym{rl}{RL}{Reinforcement ~Learning~}
\newacronym[longplural={Markov Decision Processes}]{mdp}{MDP}{Markov Decision Process}

\begin{document}

\title{Extending a Quantum Reinforcement Learning Exploration Policy with Flags to Connect Four\\
}

\author{\IEEEauthorblockN{Filipe Santos}
\IEEEauthorblockA{\textit{CISUC, DEI} \\
\textit{University of Coimbra}\\
Coimbra, Portugal \\
0000-0003-1411-3834}
\and
\IEEEauthorblockN{João Paulo Fernandes}
\IEEEauthorblockA{\textit{LIACC} \\
\textit{New York University Abu Dhabi}\\
Abu Dhabi, United Arab Emirates \\
0000-0002-1952-9460}
\and
\IEEEauthorblockN{Luis Macedo}
\IEEEauthorblockA{\textit{CISUC, DEI} \\
\textit{University of Coimbra}\\
Coimbra, Portugal \\
0000-0002-3144-0362}
}

\maketitle

\begin{abstract}
Action selection based on flags is a \gls{rl} exploration policy that improves the exploration of the state space through the use of flags, which can identify the most promising actions to take in each state. The quantum counterpart of this exploration policy further improves upon this by taking advantage of a quadratic speedup for sampling flagged actions. This approach has already been successfully employed for the game of Checkers. In this work, we describe the application of this method to the context of Connect Four, in order to study its performance in a different setting, which can lead to a better generalization of the technique. We also kept track of a metric that wasn't taken into account in previous work: the average number of iterations to obtain a flagged action. Since going second is a significant disadvantage in Connect Four, we also had the intent of exploring how this more complex scenario would impact the performance of our approach. The experiments involved training and testing classical and quantum \gls{rl} agents that played either going first or going second against a Randomized Negamax opponent. The results showed that both flagged exploration policies were clearly superior to a simple $\epsilon$-greedy policy. Furthermore, the quantum agents did in fact sample flagged actions in less iterations. Despite obtaining tagged actions more consistently, the win rates between the classical and quantum versions of the approach were identical, which could be due to the simplicity of the training scenario chosen.
\end{abstract}

\glsresetall


\section{Introduction}

In \gls{rl} \cite{sutton2018reinforcement}, we consider an agent that learns by interacting with an environment by trial and error. The policy followed by the agent dictates which actions it can take in each state, and with which probability. According to the decisions it takes, the agent receives scalar rewards based on its performance. These rewards are then used to update the agent's policy, allowing it to keep learning better actions for each situation. Since this approach does not require domain knowledge to work, it has the advantage of not needing previously labeled training data to function. As a result, \gls{rl} has been used to great success in different areas, such as games and robotics \cite{Li2018}. RL is frequently paired with Neural Networks, resulting in Deep Learning approaches \cite{Li2018}, which can work well even when dealing with large state spaces.

Quantum Computing \cite{IBM-Quantum-Learning} is a computing paradigm that takes advantage of quantum properties to achieve key advantages over classical approaches. Quantum \gls{rl} is one of the areas of this computing model, with recent research demonstrating the potential benefits of quantum approaches and quantum-inspired algorithms \cite{skolik2021quantum, wei_qier_2021}, such as needing less parameters to execute the same tasks. These advantages can also take the form of quadratic speedups and better balance between exploration and exploitation. An agent explores by selecting actions in a random manner, to find out which actions are better to select in each state. Meanwhile, when an agent chooses the action it considers optimal, it is exploiting its current knowledge. Balancing these two concepts is one of the key topics in \gls{rl}, as it leads to better training efficiency. As current quantum hardware is noisy and quantum simulations are difficult to perform for more complex systems, it is especially relevant to figure out practical use cases for Quantum Computing in the near future.

Recently, action selection taking advantage of flags \cite{Briegel2012Projective,Paparo_2014_Projective} was applied to the \gls{rl} problem of learning how to win at Checkers \cite{miguel_2021}, with the results suggesting a better training efficiency for the quantum approach, due to an improved exploration of the state space. With our work, we intend to use this exploration policy for Connect Four, to verify whether offline Deep Q-Learning can be used as a way of dealing with the complexity of the state space while using this policy, extending its applicability to a different context. Furthermore, we want to study an important metric that was not collected in that work (average number of iterations to obtain an action with a flag), as explained by its authors \cite{miguel_2021}. Since the player who goes second is at a major disadvantage in Connect Four, we also have the goal of demonstrating the impact of using this method for the more complex scenario of training as player~2. The experiments involved testing agents that trained either as player 1 or as player 2, with the classical and quantum versions of the algorithm being compared. All of the \gls{rl} agents trained by playing against an opponent that used a Randomized Negamax strategy.  

The remainder of this paper is organized as follows. In Section \ref{sec:literature}, we go over literature related with quantum \gls{rl}. In Section \ref{sec:opponent}, the opponent the agents played against is presented. Section \ref{sec:reinforcement} covers how we tackled the \gls{rl} context of Connect Four in this work, while also mentioning important differences with the approach used for Checkers \cite{miguel_2021} that our work is based on. In section \ref{sec:exploration}, we explain the exploration policies used in this work. Section \ref{sec:results} presents the setup used for the experiments, the results obtained and the discussion. Lastly, Section \ref{sec:conclusion} covers the conclusion.

\section{Literature Review}
\label{sec:literature}

We will now present some articles in the field of quantum \gls{rl}, which are especially relevant in the context of our work.

Briegel et al. \cite{Briegel2012Projective} established an \gls{rl} framework that projects an agent into simulated situations, before it chooses the action to take. Projective Simulation is centered around clips, which represent episodes stored in memory. By considering a stochastic clip network, it is possible to think about a clip being excited by a perception of the environment, which leads to the corresponding experience being replayed. That clip can then jump to a neighboring clip that contains similar experiences and so forth, through the use of classical or quantum random walks, until an actuator clip produces an action. This allows the agent to simulate possible future situations before making a choice, by projecting itself into those clips. Furthermore, tags were employed so that transitions between clips could be classified based on how they were rewarded. This way, actions that had previously been rewarded for a certain state have a higher chance of being picked again. 

Paparo et al. \cite{Paparo_2014_Projective} expanded upon the initial concept of Projective Simulation by further exploring a specific variant of this model, known as Reflecting Projective Simulation. This name derives from the fact that the agent reflects upon its choices when it has to select an action. It does so by executing a random walk through the clip network and then checking if a tagged action was obtained. The quantum version of the random walk over the clip network mimics the search for marked elements seen in \cite{Magniez2011searchViaQuantumWalk}, with the marked elements being the flagged actions. If the action was tagged, then it is selected, otherwise the process repeats up to a predefined number of tries, which corresponds to the agent reflecting on that choice. Their work proved that it is possible to achieve a quadratic speedup with the quantum version of the framework, while still respecting that the actions need to be chosen according to a re-normalized probability distribution.

Teixeira et al. \cite{miguel_2021} applied quantum \gls{rl} to the game of Checkers, by using a quantum-enhanced agent. In order to avoid dealing with a large state space composed of every possible board position, they instead represented the problem using the relative position of the pieces on the board and their number. This way, similar board states were interpreted similarly by the agent. As for the agent itself, it took advantage of a quantum flag update mechanism inspired by \cite{Paparo_2014_Projective}. 

These works all increasingly show the potential of using tags for a quantum \gls{rl} approach, which is the strategy we chose to use for our exploration policies in Section \ref{sec:exploration}.

\section{A Randomized Negamax Opponent}
\label{sec:opponent}

The \gls{rl} agents were all trained and tested against a Randomized Negamax opponent, so as to have a fair comparison between them. When we have a zero-sum game \cite{russell2010artificial}, it is possible to use a Minimax \cite{russell2010artificial} strategy, with Negamax being one of its variants. The key concept behind this approach is to look ahead a certain number of moves in a recursive manner. In each iteration, the states that result from each possible action are evaluated, with the evaluation for the final states being achieved through an evaluation function, which will be explained in more detail at the end of this section. The action that maximizes a player's score is then picked, which corresponds to minimizing the other player's score. Essentially, this leads to choosing the best move according to an evaluation function, while assuming that the other player is attempting to do the same. Specifically for Negamax, each player considers the negated score from the other player, as seen in \eqref{eq:negamax}, in which $A$ and $B$ represent the two players and $depth$ is the number of moves being looked ahead.

\begin{equation} \label{eq:negamax}
    score_A = -Negamax(depth-1)_B
\end{equation}

As for the $depth$ parameter, a value of 2 was selected. Using a $depth$ of 1 would mean that the opponent only takes into account the positions it can reach after making its move, while completely ignoring what the agent is capable of doing during the action selection that comes after that. As for higher depths, while they would make for interesting experiments, preliminary tests revealed that even a depth of 4 would lead to a very significant increase in the amount of time necessary to run a set of games. This longer play time results from two factors. Firstly, the opponent takes longer to make a decision, as it needs to look further ahead for every choice it makes. Secondly, since the opponent is effectively a stronger player, the agents are forced to train for a much higher number of episodes, so that they can achieve reasonable win rates. Due to time constraints, doing multiple runs of these longer experiments was simply not feasible.

The randomized version of the algorithm we used differs from the standard approach in that it has a random probability $\omega = 0.3$ of choosing a random action, when selecting the action to execute after returning from all the recursive calls. This random selection gives priority to actions with positive scores and is only applied in non-critical move choices. For instance, if the opponent has a guaranteed winning move, it will always be chosen. The reason behind adding this randomness to the algorithm, along with its specific mechanisms, was to create a more difficult challenge for the \gls{rl} agents, stopping them from winning in overly trivial ways, such as simply stacking a specific column with discs. The value $\omega = 0.3$ was picked after some preliminary tests with different values for $\omega$. Values lower than this led to the agents playing too many similar games, which defeated the purpose of using a randomized approach. Meanwhile, values higher than 0.3 were making the challenge too trivial for the agents, as the use of too many random moves made it simple to setup winning scenarios.

This Randomized Negamax algorithm makes use of Alpha-Beta Pruning \cite{russell2010artificial} in order to explore the game tree more efficiently. This technique relies on parameters $\alpha$ and $\beta$, which are lower and upper bounds on the evaluations, respectively. When a score bigger than the upper bound is found for player $A$, the analysis of the corresponding tree branch can stop, as player $B$ would pick a move that denies the opportunity of surpassing that upper bound. The same logic applies when the players are reversed. This way, it is possible to reduce the amount of time spent evaluating actions from the game tree.

The evaluation function used to evaluate the current state is an heuristic, with the state in the algorithm being a matrix that represents the Connect Four grid, which has 7 columns and 6 rows. The idea is to consider every 4 by 4 square inside the Connect Four grid, with the tags for the players 1 and 2 being identified with 1 and -1, respectively. By summing along the column, row and both diagonals of each of those squares, we obtain four numbers, which represent the advantage a player has in each of those four lines. Positive values indicate that player 1 has the advantage, while negative values signify the opposite. We then consider the highest and lowest of those four values. If the highest value returned is higher than 1, then we sum the double of its value to a $max\_heuristic$ variable, with an identical strategy being applied for the lowest value. The reason for only considering values higher than 1 and lower than -1 is to exclusively benefit players that are connecting multiple discs, while penalizing lines that have discs from the other player in the middle. The values are multiplied by 2 to create a bigger gap between, for instance, connecting 2 and 3 discs. The final value returned is the sum of the $max\_heuristic$, which is always positive when not 0, with the $min\_heuristic$, which is always negative when not 0. This value is negated if the player being evaluated is player 2, as that player's discs have the value -1. 
\section{Reinforcement Learning for Connect Four}
\label{sec:reinforcement}

In this section, we define Connect Four as a \gls{mdp} and explain how we dealt with the complexity of the state space using an offline Deep Q-learning strategy. We also talk about key differences with the Q-Learning approach used for Checkers \cite{miguel_2021}, which this work is based on. 

\subsection{Connect Four as a Markov Decision Process}

We can define the \gls{rl} problem of Connect Four as an \gls{mdp}, as it obeys the property of the state of the system summarizing all the relevant information about the sequence of states that led to the current one, while being enough to fully determine the future of the game \cite{sutton2018reinforcement}. We can then define this \gls{rl} problem using the tuple ($S$, $A$, $P$, $R$, $\gamma$) \cite{Li2018}.

The state space $S$ is defined by all the positions that can be reached by the players, which correspond to every possible combination of empty spaces and discs from either player. As the board has 7 columns and 6 rows, there are $7 \times 6=42$ spaces on the grid, which means we have at most \(42^3\) possible states (since each space can be empty, have a piece from player A or have a piece from player B). By ignoring positions that are impossible to reach, such as having discs in the middle with nothing below them, we can get a lower state space complexity of $4.53 \times 10^{12}$ \cite{Edelkamp2008}. In the context of this work, the state space is represented through a matrix with 7 columns and 6 rows, which corresponds to the Connect Four grid. In this matrix, $0$ represents an empty space, $1$ represents a disc from player 1 and $-1$ represents a disc from player 2.

As for the action space $A$, it is defined by the columns in which discs can be inserted. With the board having seven columns, the action space consists of seven minus the number of full columns in which discs cannot be inserted. The reward function $R$ was adapted from \cite{wisney2019} and is shown in \eqref{eq:reward_function}, in which we consider the reward for taking action $a$ in state $s$ and reaching state $s'$. A discount factor $\gamma$ of 1 was chosen for this problem, which means the expected return is taking into account the sum of rewards without discounting them \cite{sutton2018reinforcement}.

\begin{equation} \label{eq:reward_function}
  R(s, a, s') =
    \begin{cases}
      1 & \text{if the agent wins the game}\\
      0.5 & \text{if the game is a draw}\\
      -1 & \text{if the agent loses the game}\\
      0 & \text{if the winner is yet to be decided}
    \end{cases}       
\end{equation}

\subsection{Dealing with the State Space Complexity}

In order to deal with the previously mentioned state space complexity of the problem, an offline Deep Q-Learning approach was utilized, using as a starting point the work from \cite{wisney2019}, in which classical offline Deep Q-Learning was used to train Connect Four \gls{rl} agents. Preliminary tests revealed this offline Deep strategy to be more effective for this problem than the ones that had been used for Checkers \cite{miguel_2021}. Most notably, that work featured online non-Deep Q-Learning and the Checkers board was split into different sections for the state representation, with each state identified by how many pieces were in each one. This way, similar states were grouped together, which allowed them to deal with the high dimensionality of the state space of that board game. Since the state space complexity of Connect Four is not as high and the Neural Network used already allows it to generalize past knowledge, we decided to represent the state with a simple matrix that corresponds to the Connect Four board, as explained in the previous subsection. This shows that there are different ways of dealing with the problems of dimensionality when applying action selection based on flags, some of which may be more effective for certain problems than others. 

It is also worth noting that unlike \cite{miguel_2021}, we did not utilize reward shaping. Reward shaping is used to deal with the fact that the winner of a game is only decided during the last transition of the game, which means that this would typically be the only transition with a non-zero reward. In other words, they introduced smaller rewards to every transition, based on whether the agent was achieving certain goals that in theory should give it a better winning probability. In our work, some preliminary tests were done with this type of approach. However, the agents trained with this methodology were not learning how to win efficiently, so we decided to drop it. It is possible that better reward shaping functions could have been used, but this shows that this type of approach is not strictly necessary and might not be ideal for every board game. It is worth considering that the offline Deep Learning approach employed helped the agents to get a better grasp of the impact of each transition, due to it training after each batch of 300 games, using the Neural Network to evaluate every position from the batch, ultimately attributing similar $Q$ values to similar state-action pairs. 

As for Q-Learning, it is a model-free, Temporal-Difference learning method \cite{sutton2018reinforcement}. The aim of this technique is to try to learn the optimal action-value function $Q$ directly, with $Q(S_t,A_t)$ representing the expected future reward for the state-action pair \cite{sutton2018reinforcement}. Q-Learning can be defined by \eqref{eq:Q_Learning}, in which $\alpha$ is the learning rate, $R_{t+1}$ is the reward for selecting action $A_t$ in state $S_t$, $\gamma$ is the previously mentioned discount factor and $\underset{a}{max} Q(S_{t+1},a)$ is the highest action value $Q$ out of the possible transitions for the following state. It is also worth mentioning that $R_{t+1} + \gamma \underset{a}{max} Q(S_{t+1},a)$ is considered the target value, while $R_{t+1} + \gamma \underset{a}{max} Q(S_{t+1},a) - Q(S_t,A_t)$ is considered the Temporal Difference error. We can think of an optimal action as selecting the action with the highest $Q$ value for a state. The learning rate $\alpha$ was set to 0.8 in this work.

\begin{multline} \label{eq:Q_Learning}
    Q(S_t,A_t) \leftarrow Q(S_t,A_t) + \alpha[R_{t+1} + \gamma \underset{a}{max} Q(S_{t+1},a)\\
    - Q(S_t,A_t)]
\end{multline}

The approach employed is considered offline since the agents played a certain number of games, equal to the size of the batch (which was fixed at 300 episodes) and were then trained on that batch before moving on to the next batch of games. Each batch ran for 5 epochs, which means that the Neural Network spent 5 cycles with the same batch. This is the same way that the agents were trained in \cite{wisney2019}. 

A Neural Network identical to the one from \cite{wisney2019} was used for the Deep Learning of the Q-function, in order to learn the $Q$ values at the end of every batch, allowing \eqref{eq:Q_Learning} to be solved for each transition stored in the batch. For each state-action pair, the network predicts its value by considering the Connect Four matrix which represents the corresponding afterstate. An afterstate \cite{sutton2018reinforcement} is the state we reach after taking the chosen action, but before the opponent makes their choice, as that is unknown information. We can use this representation since in Connect Four we know how the state will look like after we choose our move. The only thing we don't know is how the opponent will reply. Using this representation is considered advantageous for multiple reasons, such as grouping together states that can be reached by different state-action pairs \cite{sutton2018reinforcement}. The only differences in our Neural Network are the following two: the input has a shape of (6, 7) instead of (7, 7), as we directly consider the afterstate as the input; the first convolutional layer has a convolution window of (4, 4) instead of (5, 5).

\section{The Exploration Policies}
\label{sec:exploration}

A common problem in \gls{rl} is balancing exploration with exploitation. An agent explores by trying new actions (usually randomly) and exploits by picking an action considered optimal by its current policy. If an agent does not explore enough, it will be missing better moves that it simply never tried to use. However, if an agent does not exploit enough, than its behavior will be almost random and it will not take enough advantage of what it is learning.

One of the approaches tested to deal with this issue was $\epsilon$-greedy. The idea is to pick an exploratory action with probability $\epsilon$ and to choose an optimal action with 1-$\epsilon$ probability. This is a very generic strategy that can be applied to most \gls{rl} problems in order to achieve some decent results. An equation describing it can be seen in \eqref{eq:epsilon-greedy}. The $\epsilon$ that defines this approach is updated following \eqref{eq:epsilon}, in which the $+ 1$ in the denominator is used to avoid $log(1) = 0$ when $episode = 1$, as that would lead to $1/0$.

\begin{equation} \label{eq:epsilon-greedy}
  \epsilon-greedy =
    \begin{cases}
      \epsilon & \text{Pick a random action}\\
      1-\epsilon & \text{Pick an optimal action}
    \end{cases}       
\end{equation}%

\begin{equation} \label{eq:epsilon}
    \epsilon_{episode} = \frac{1}{log(episode + 1)}
\end{equation}%

The following subsection covers a more complex exploration policy, which is also the basis of the quantum exploration policy employed.

\subsection{Classical Flagged Action Selection}

In order to achieve a better exploration of the state space, we can instead employ action selection based on flags (which we will interchangeably refer to as tags), as seen in \cite{miguel_2021}. This strategy is also used in the quantum version of the algorithm, as will be explained in the following subsection. We will cover both the key concepts related to this approach, as well as how they are applied in this work.

The idea to use flags for action selection was initially presented in \cite{Briegel2012Projective}. Each action from each state can either have a flag or not, which identify which ones have been rewarded positively in previous action selections. By making use of this mechanism while selecting an action, the agent can reflect on its choice, so that it is more likely to select an action capable of yielding a good result, allowing it to more efficiently explore the state space. This process is known as reflection.

\begin{figure}[b]
    \centering
    \includegraphics[scale=0.6]{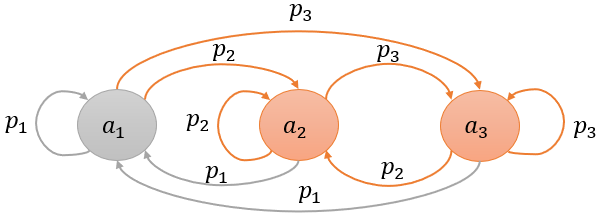}
    \caption{A Markov chain for a specific state, with the transition probability to each action matching the probability of selecting that action when in that state. The flagged actions are colored orange.}
    \label{fig:Markov_Chain}
\end{figure}

In \cite{Paparo_2014_Projective}, a specific way to utilize the flags in the reflection process is mentioned. Though there are some differences, as we will see later, this strategy is the baseline for the approach utilized in this work, so we will start by explaining how it works. Every action of each state starts out by being tagged. When an action is selected for a certain state, if the corresponding reward is not positive, then the tag is removed. If there is a situation in which all the actions relative to a certain state have lost their flags, then all of them become flagged once again, ensuring that there is always at least one action with a flag for each state. During the reflection process, the agent repeatedly samples random actions for a predefined number of iterations $R$, attempting to obtain one that has a flag, which corresponds to reflecting about its choice. Each of these steps of deliberation can be seen as a classical random walk over a directed weighted graph, in which the transition probabilities lead the agent from the state to each of the possible actions. More accurately, each step can be described as a classical random walk over a Markov Chain, with the transition probabilities from one action to another having the same value as the probability of selecting that action while in that state. The reflection process can be thought of as a random walk over this Markov Chain, repeated for up to $R$ iterations, which can be visualized in the example illustration from Fig. \ref{fig:Markov_Chain}. The objective of this iterative process is to ultimately approximate $\overset{\sim}{\pi}_s$ for each state $s$, which is a re-normalized version of the original stationary distribution $\pi_s$, but with support only over the flagged actions $f(s)$. $\overset{\sim}{\pi}_s$ can be visualized in \eqref{eq:re_normalized_distribution}.

\begin{equation} \label{eq:re_normalized_distribution}
    \overset{\sim}{\pi}_s(a) = 
        \begin{cases}
            \frac{\pi_s(a)}{\sum_{a' \in f(s)} \pi_s(a')} & \text{If a} \in f(s) \\
            0 & \text{Else}
        \end{cases}
\end{equation}%

\begin{figure}[b]
    \centering
    \includegraphics[scale=0.5]{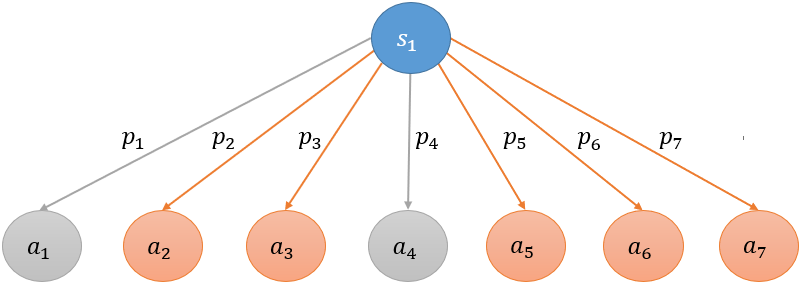}
    \caption{A graph representing the transition probabilities for each action of a specific state, with each action corresponding to inserting a disc on that column. The flagged actions are colored orange.}
    \label{fig:Transition_Probabilities}
\end{figure}

For Connect Four specifically, we consider the transition from each state into its available actions, each of which corresponds to inserting a disc on a specific column. These actions all start out with a tag, which may be removed by a process that will be explained later in this subsection. We can see these transitions illustrated in Fig. \ref{fig:Transition_Probabilities}. Each iteration of the reflection process can be seen as a random walk on a Markov Chain similar to the one on Fig. \ref{fig:Markov_Chain}, with the transition probabilities matching the ones seen here.

This strategy of gradually removing flags according to a criteria and deliberating on the action choice for a specific number of iterations \cite{Paparo_2014_Projective} was used as a starting point in \cite{miguel_2021}, in conjunction with a soft-max policy. The objective of this approach was to improve the exploration of the state space for Checkers. This strategy is used as a basis for our work and is detailed below.

The main difference between this method and the previously mentioned one \cite{Paparo_2014_Projective} is in the update mechanism of the flags. Instead of deleting flags based on whether or not their transitions were rewarded, instead they are removed if their corresponding $Q$ value was smaller than $0$. That way, the update takes into account a more nuanced view of how each transition impacts the odds of winning a game. This is due to the fact that simply using the immediate reward leads to ignoring that a move evaluated as worse at the moment can potentially lead to a game winning scenario and vice versa. In general, using the $Q$ value makes the most sense when considering more complex environments, in which the immediate reward does not necessarily present the full picture. This change is especially crucial for our Connect Four approach, considering that the reward function only rewards the final transition. Meanwhile, if the action selected had a positive $Q$ value despite not having a tag, then it receives a new one. Also, when all actions of a certain state lose their flags, all of them except the last one picked gain it back, as opposed to simply allowing every action of that state to have a flag again. This mechanism for updating flags happens after the reflection process, in which the agent attempts to sample a flagged action for a maximum of $R$ iterations, as previously explained. During the reflection process, if no flagged action is obtained in $R$ steps, then the last action sampled is chosen. Keep in mind that this is the same process proposed in \cite{miguel_2021}. For our work, the maximum number of iterations $R$ was set to 5.

As for the probabilities of selecting each action for each state, a Boltzmann distribution, also known as soft-max policy, was used. The key concept for this distribution is to make the probability of selecting an action related to its $Q$ value. That way, action selection can benefit from randomness in order to improve the exploration, while still exploiting its current knowledge by choosing actions with higher $Q$ values more often. Furthermore, it makes use of a temperature $T$, which controls the balance between exploration and exploitation. $T$ decreases with the number of episodes, which corresponds to gradually increasing the relative probability of selecting actions with high $Q$ values, leading to more exploitation as the agent becomes more familiar with the environment. This probability distribution can be seen in \eqref{eq:boltzmann}, in which $A(s)$ represents the set of actions for state $s$, $a$ represents an action and $Q(s, a)$ represents the $Q$ value of a state-action pair.

\begin{equation} \label{eq:boltzmann}
    P(a|s) = \frac{ e^{Q(s,a)/T} }{ \sum_{a' \in A(s)} e^{Q(s, a')/T} }
\end{equation}

\begin{figure}[t]
    \centering
    \includegraphics[scale=0.3]{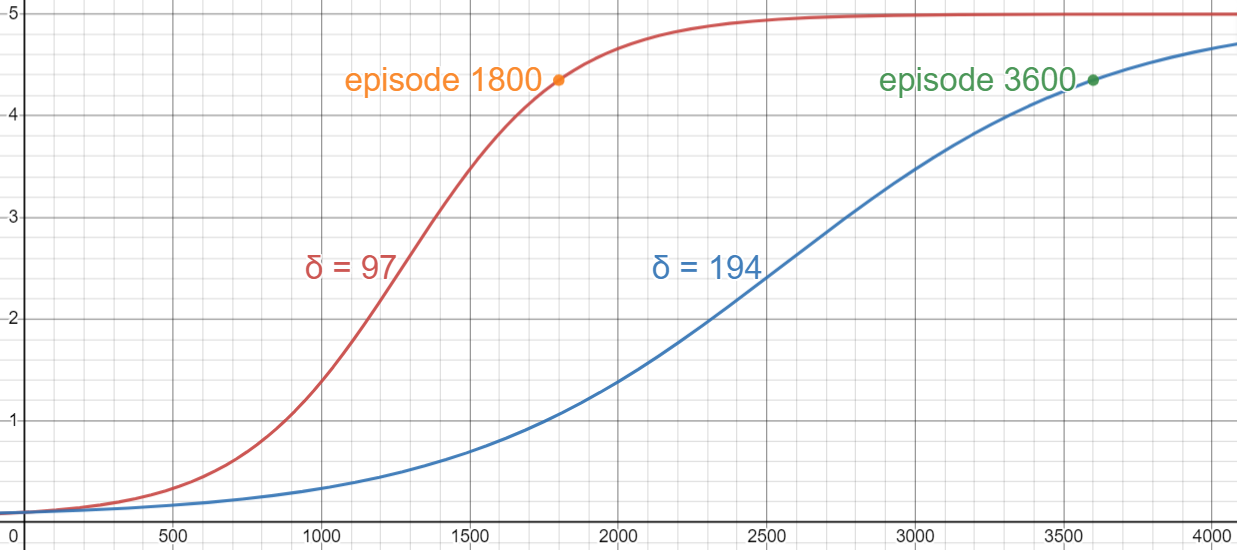}
    \caption{The Inverse Temperature Functions.}
    \label{fig:temperature}
\end{figure}

As for the temperature parameter $T$ seen in \eqref{eq:temperature}, since the idea of using a Boltzmann distribution was inspired by \cite{miguel_2021}, the temperature function used in that work was used as a basis, but altered to fit the context of our approach. In particular, a parameter that we refer to as $\delta$ was changed so that a high inverse temperature $\frac{1}{T}$ is reached around 1800 or 3600 episodes, depending on whether it is a going first or a going second agent. In Fig. \ref{fig:temperature}, the inverse temperature functions can be observed.

\begin{equation} \label{eq:temperature}
    T = 0.2 + \frac{20-0.2}{1 + e^{0.35 \times ( episode / \delta ) }} 
\end{equation}

\subsection{Quantum Flagged Action Selection}

The quantum version of the action selection based on tags was also introduced in \cite{Paparo_2014_Projective}. This method was used in tandem with a  coherent controlization \cite{Dunjko_2015} encoding strategy in \cite{miguel_2021}, which is the approach we chose to utilize for our work.

As the objective of this type of exploration policy is to output flagged actions with high probability, it is crucial to guarantee that this happens consistently. This is where Quantum Computing can come into play, making it possible to achieve a quadratic speedup in the number of iterations necessary to obtain a tagged action during the reflection process \cite{Paparo_2014_Projective}, as will be explained during this subsection. Achieving this kind of speedup can be very beneficial, as it implies that we are not reaching the maximum number of iterations $R$ as often, which leads to random actions being sampled more rarely and flagged actions being obtained more consistently. This way, the agent takes better advantage of the reflection process, leading to a better balance between exploration and exploitation.

The key difference with the classical version is that in each step of the iterative process, a quantum walk over a Markov Chain is used to approximate the $\overset{\sim}{\pi}_s$ seen in \eqref{eq:re_normalized_distribution}, as opposed to its classical counterpart \cite{Paparo_2014_Projective}. This process is similar to the search via quantum walk presented in \cite{Magniez2011searchViaQuantumWalk}, in which a randomized Grover search is applied in order to find marked elements. In this case, the marked elements correspond to the flagged actions. 

The first step of this quantum reflection process is to encode the stationary distribution $\pi_s$, which leads us to the quantum state $\ket{\pi_s}$. This state can be seen in \eqref{eq:quantum_distribution}, with $\pi_s(a)$ representing the probability of selecting action $a$ while in state $s$ and $\ket{a}$ being its corresponding basis state. This part of the process will be explained in more detail in the following subsection, in which the encoding is covered.

\begin{equation} \label{eq:quantum_distribution}
    \ket{\pi_s} = \sum_{a \in A(s)} \sqrt{ \pi_s(a) } \ket{a}
\end{equation}

From there, it is necessary to repeatedly apply two reflection operators in succession, in an identical manner to a randomized Grover algorithm. Note that the reflections performed by the reflection operators are related with the quantum states, not to be confused with the reflection process in which the agent reflects over its actions for a maximum of $R$ steps. The number of repetitions for the two reflections, $m$, is chosen uniformly at random in $[0, 1/\sqrt{\epsilon}]$, with $\epsilon$ representing a lower bound on the probability of an action having a flag. The reason for this choice is due to the fact that a randomized Grover algorithm should be applied a number of times chosen uniformly at random from that interval \cite{Magniez2011searchViaQuantumWalk}.

The first operator is a reflection over the flagged actions, which is referred to as $ref(f(s))$. Its aim is to check which actions have a tag and then flip the phase of every action belonging to $f(s)$. This is achieved with the use of a Boolean function $F(a)$, which returns 1 if the action has a tag and 0 otherwise. By applying the oracle $-1^{F(a)}$, the actions are evaluated, with the flagged ones being subjected to a phase kickback, which is a rotation by a phase of $\pi$ radians. Notice that this is identical to the first step of Grover's algorithm. This operator is represented in \eqref{eq:ref(f(s))}.

\begin{equation} \label{eq:ref(f(s))}
    \ket{a} \rightarrow 
        \begin{cases}
            -\ket{a} & \text{If a $\in f(s)$}\\
            \ket{a} & \text{Else}
        \end{cases}
\end{equation}%

As for the second operator, it is a reflection over the encoded stationary distribution $\ket{\pi_s}$. By reflecting the actions over $\ket{\pi_s}$ after having flipped the phase of the ones with flags, the amplitudes of the flagged actions will greatly increase, while the amplitudes of the other actions will slightly decrease. This is identical to the inversion around the mean in Grover's algorithm, known as Grover's Diffusion Operator. This operator can be seen in \eqref{eq:diffuse}, with $I$ representing the identity. 

\begin{equation} \label{eq:diffuse}
    ref(\pi_s) = 2 \ket{\pi_s} \bra{\pi_s} - I
\end{equation}

Since we are considering the transition matrix of a rank-one Markov Chain \cite{Paparo_2014_Projective}, it is possible to obtain this operator using the transformations shown in \eqref{eq:diffuse2}, as explained in \cite{Dunjko_2015}. $D_0$ represents a reflection over the state $\ket{0}$. $U$ is the operator used to encode the stationary distribution and will be explained in more detail in the following subsection. $U^\dagger$ is the conjugate transpose of $U$. First, $U^\dagger$ reverses the encoding, then $D_0$ rotates over state $\ket{0}$ and lastly $U$ encodes the probability distribution once again. This process is equivalent to a reflection over $\pi_s$.

\begin{equation} \label{eq:diffuse2}
    ref(\pi_s) = U D_0 U^\dagger
\end{equation}

Finally, the quantum circuit is measured. If the action obtained is tagged or the maximum number of reflections $R$ has been reached, then it is selected by the agent. Otherwise, the entire process repeats for that state.

As the average number of repetitions required for this process is $1/\sqrt{\epsilon}$, as opposed to $1/\epsilon$ for the classical case,  we achieve a quadratic speedup for obtaining a flagged action \cite{Magniez2011searchViaQuantumWalk}, while still obeying the fact that the actions need to be sampled according to the re-normalized distribution \cite{Paparo_2014_Projective}. While this strategy is not regarded as being optimal for simple searching problems \cite{Magniez2011searchViaQuantumWalk}, it has been shown to be optimal when it is merely necessary to output an action from a good approximation of the re-normalized distribution $\overset{\sim}{\pi}_s$ \cite{Paparo_2014_Projective}. 

\subsection{Encoding The Action Selection Probabilities}

The encoding of the action selection probabilities was done taking advantage of the nested coherent controlization scheme introduced in \cite{Dunjko_2015}, using the code from \cite{miguel_2021}, which also employed this approach for the encoding, as mentioned before. Coherent controlization is built around a unitary operator $U$, composed of a single-qubit Y rotation to which other controlled Y rotations are added in a nested process. 

This type of encoding utilizes angles as parameters, with these corresponding to the probabilities from the stationary distribution, which in this work is the Boltzmann distribution from the previous subsection. As such, it is necessary to convert the classical probabilities into angles. First, we consider the relation $p_1 = cos^{2}(\theta_1/2)$, in which $p_1$ is the sum of the probabilities of half the actions and $p_2$ represents the sum of the other half. For Connect Four $p1$ typically corresponds to the probabilities of four actions, $p_1=p''_1+p''_2+p''_3+p''_4$, as there are seven actions available in the most common scenario, in which no column is full. The first angle is then obtained by $\theta_1 = 2 \times arccos(\sqrt{p_1})$ and a Y rotation $U_Y(\theta_1)$ is applied on the first qubit. Then, two controlled operations are applied on the second qubit, in which we consider that half of the actions go to one operation and the other half to the other. These are conditioned on that first qubit, with $U_Y(\theta_2)$ being applied if it is in state $\ket{0}$, while $U_Y(\theta_3)$ is applied if it is in state $\ket{1}$. The angles for these rotations are then based on the re-normalized probabilities for each side, with $\theta_2 = 2 \times arccos(\sqrt{(p''_1+p''_2)/p_1})$ and $\theta_3 = 2 \times arccos(\sqrt{(p''_5+p''_6)/p_2})$. Both of these operations are then followed by two controlled Y rotations applied on qubit 3 and conditioned on qubit 2, which work the same way as the two controlled operations we just saw. In other words, we once again divide the actions across the operations, this time ending up with four new Y rotations ($U_Y(\theta_4)$, $U_Y(\theta_5)$), $U_Y(\theta_6)$, $U_Y(\theta_7)$, each one with an angle that depends on two actions. For example, the first one of these would be $\theta_4 = 2 \times arccos(\sqrt{p''_1/(p''_1+p''_2)})$.

The process described ends at that point for Connect Four, as it only has seven actions and their probabilities were divided as much as possible. However, it is possible to apply it to contexts in which there are more actions, by further adding coherent control with this nested scheme. For an illustration, see Fig. 3 from \cite{Dunjko_2015}.
\section{Experimental Setup, Results and Discussion}
\label{sec:results}

The player 1 agents were trained by playing for 1800 episodes against the Randomized Negamax opponent, while the player 2 agents were trained for 3600 episodes instead. This choice was due to the fact that going second is a considerable disadvantage in Connect Four, leading to a more complex scenario. The number of states explored was recorded, so we can understand how the win rates are impacted by this factor.

For the agents with action selection based on tags, the average number of iterations to obtain a flagged action was registered, which is a metric that was missing in the work for Checkers \cite{miguel_2021}, as mentioned by its authors. This metric is used to compare the classical and quantum versions of this approach on how consistent they are at sampling flagged actions. Specifically, a lower average number of iterations indicates that flagged actions are obtained more frequently, since the reflection limit $R$ is not being hit as often, which would result in the last action sampled being chosen instead. Note that when a flagged action is not hit during action selection, surpassing the limit $R$, the corresponding number used for the average will keep summing the iterations of the following action selection process, and so on, until a flagged action is eventually hit. For the quantum agents, Qiskit’s Aer simulator was used as the backend.

After training, the agents were tested against the opponent for 1000 episodes, with the number of wins being registered. The objective of these tests was to obtain a good representation of the performance of the agents after their training had been completed. Note that while testing, all of the agents made use of an optimal action selection function, which always returns the action with the highest $Q$ value. This was done with the intent of enabling a fair comparison between the different agents, only taking into account the actions considered optimal by their current policy, which corresponds to exploiting the knowledge they obtained while training.

All of the results obtained were averaged over 20 independent runs, each with a unique seed. We also took into account the corresponding standard deviation for these values.

Table \ref{tab:player1Agents} shows the results for the player 1 agents. The classical $\epsilon$-greedy agents were exposed to a significantly higher number of states, which could be due to them overly favoring exploration in favor of exploitation. This higher number of states explored could also be a consequence of them having played longer games, which would naturally lead to exploring more states. 

\begin{table}[b]
\centering
\caption{Player 1 Agents vs Randomized Negamax}\label{tab:player1Agents}
\begin{tabular}{|c|c|c|c|}
\hline
Agent & Iterations & States & Win \%\\
\hline
Classical, $\epsilon$-greedy &   & $\textbf{9515} \pm \textbf{2405}$ & $\textbf{51.8} \pm \textbf{28.4}$\\
Classical, tags &  $1.394 \pm 0.140$ & $8368 \pm 474$ & $86.5 \pm 4.9$\\
Quantum, tags & $\textbf{1.345} \pm \textbf{0.103}$ & $8253 \pm 386$ & $85.5 \pm 4.6$\\
\hline
\end{tabular}
\end{table}

Looking at the win rates, we can conclude that the tagged action selection policies are much more effective at training the agents than $\epsilon$-greedy, as they both managed to achieve reasonable results, while the classical $\epsilon$-greedy agents had a very poor performance. We also observe that the $\epsilon$-greedy policy led to an unusually high standard deviation for the states explored and the win rates. A closer look at the individual results for each run revealed a big variation in performance, with half of them achieving win rates of $61.6$ - $88.0\%$ and the other half only reaching $13.6-36.0\%$. Due to the inconsistent results obtained, it is fair to say that this approach was likely highly reliant on luck when it came to exploring the right states, as opposed to the flagged approaches, which were significantly more consistent. 

For the exploration policies based on tags, the quantum agents obtained flagged actions in less iterations, which is in line with the theory surrounding them. While the difference might seem small, note that the maximum number of iterations $R$ was set to $5$ and that a considerable number of flagged actions are obtained in the first iteration. As such, it becomes clear that a difference of $0.049$ is still substantial. Furthermore, the standard deviation for this metric was much lower for the quantum agents, which represents a higher level of consistency. However, the win rates were very similar between both approaches, which shows that the advantage of sampling flagged actions more often may have had no impact in this scenario.

We can see the results of the player 2 agents in Table \ref{tab:player2Agents}. The quantum agents once again obtained flagged actions in less iterations and had a much lower standard deviation. That being said, the win rates are identical between both versions of the exploration policy, just as they were in the player 1 setting. 

The results from both tables seem to show that obtaining tagged actions more frequently did not seem to have an influence on the chances of winning a game. However, it is possible that these scenarios were too simple for the agents to solve. Had they have been more complex, it is possible that more consistently sampling flagged actions would have had a positive impact. This is a potential aspect to explore in future work, which could involve playing against opponents that can look ahead more than 2 moves.

\begin{table}[b]
\centering
\caption{Player 2 Agents vs Randomized Negamax}\label{tab:player2Agents}
\begin{tabular}{|c|c|c|c|}
\hline
Agent & Iterations & States & Win \%\\
\hline
Classical, tags &  $1.474 \pm 0.140$ & $13210 \pm 658$ & $69.9 \pm 2.9$\\
Quantum, tags & $\textbf{1.396} \pm \textbf{0.065}$ & $13379 \pm 834$ & $70.6 \pm 2.3$\\
\hline
\end{tabular}
\end{table}
\section{Conclusion}
\label{sec:conclusion}

In this work, we adapted Quantum Flagged Action Selection to Connect Four. The objective was to better balance how much an agent should explore and how much it should exploit, by taking advantage of a quadratic speedup for outputting flagged actions. Doing so extended the scope of this technique to board games other than Checkers, showcasing its general usefulness in this domain. 

In order to test this technique, we created a Randomized Negamax opponent for the agents to play against, whose strategy is similar to a standard Negamax, but with some added randomness in its move choices. This way, the agents had a more complicated challenge to overcome and were all tested in equal conditions. To deal with the complexity of the state space, an offline Deep Q-Learning approach was employed, showing that it is also possible to deal with this issue in different ways than previously explored. 

The results obtained show that the exploration policies based on tags were superior to a simple $\epsilon$-greedy strategy. We also found that the quantum agents sampled flagged actions in less iterations, which in theory should lead to them better exploring the state space. However, the win rates themselves were fairly close, which does not line up with this supposedly improved exploration. It is possible that the scenarios presented to the agents were not complex enough to demonstrate this advantage. That being said, sampling flagged actions in fewer iterations is still an advantage by itself, with a potential impact for active learning scenarios.

For future work, it would be possible to scale the problem by using more complex opponents. It would also be interesting to verify if this approach can be applied to a more complex environment, with Chess being the most natural choice, as it has a much higher complexity than both Connect Four and Checkers and is a well known and studied game.

\bibliographystyle{IEEEtran}
\bibliography{main.bib}

\end{document}